\documentclass{svproc}
\usepackage{microtype}
\usepackage{setspace}
\usepackage{subcaption}
\usepackage{listings}
\usepackage{graphicx}
\usepackage{amsmath,amssymb}
\usepackage{booktabs,multirow}
\usepackage{array}
\usepackage{xcolor}
\usepackage{url}
\usepackage{enumitem}
\usepackage[square,comma,numbers,sort&compress,sectionbib]{natbib}
\setlist{nosep}
\usepackage{caption}
\usepackage{subcaption}

\usepackage{tikz}

\usetikzlibrary{shapes, arrows, positioning, arrows.meta, calc}
\tikzset{
  arrow/.style={-Stealth, thick}
}

\newcommand{\averitec}{AVeriTeC}

\begin{document}
\mainmatter

\title{One LLM to Train Them All: Multi-Task Learning Framework for Fact-Checking}

\titlerunning{MTL for Fact-Checking}  %
\author{Malin Astrid Larsson \and Harald Fosen Grunnaleite \and Vinay Setty}
\authorrunning{Larsson \and Grunnaleite \and Setty}

\institute{
Factiverse and University of Stavanger, Stavanger, Norway\\
\email{\{vinay.j.setty,malin.larsson,harald.grunnaleite\}@uis.no}
}

\authorrunning{Larsson et al.} %

\maketitle

\begin{abstract}
Large language models (LLMs) are reshaping automated fact-checking (AFC) by enabling unified, end-to-end verification pipelines rather than isolated components. While large proprietary models achieve strong performance, their closed weights, complexity, and high costs limit sustainability. Fine-tuning smaller open weight models for individual AFC tasks can help but requires multiple specialized models resulting in high costs. We propose \textbf{multi-task learning (MTL)} as a more efficient alternative that fine-tunes a single model to perform claim detection, evidence ranking, and stance detection jointly. Using small decoder-only LLMs (e.g., Qwen3-4b), we explore three MTL strategies: classification heads, causal language modeling heads, and instruction-tuning, and evaluate them across model sizes, task orders, and standard non-LLM baselines. While multitask models do not universally surpass single-task baselines, they yield substantial improvements, achieving up to \textbf{44\%}, \textbf{54\%}, and \textbf{31\%} relative gains for claim detection, evidence re-ranking, and stance detection, respectively, over zero-/few-shot settings. Finally, we also provide practical, empirically grounded guidelines to help practitioners apply MTL with LLMs for automated fact-checking. 
\end{abstract}

\section{Introduction}

The rapid spread of misinformation across social and online media continues to affect public trust,  democratic discourse, and public health. Human fact-checking remains essential but cannot scale to the velocity and volume of online content, motivating the development of \emph{automated fact-checking} (AFC) systems that integrate language understanding, retrieval, and reasoning \cite{Guo:2022:TACL}. Modern AFC pipelines comprise three core components: \textit{claim detection}, identifying check-worthy statements \cite{Hassan:2017:KDD,BarronCedeno:2020:arXiv}; \textit{evidence retrieval and re-ranking}, selecting most relevant passages~\cite{Maillard:2022:arXiv,Schlichtkrull:2023:NeurIPS,V:2025:TheWebConf}; and \textit{stance or veracity prediction}, assessing whether the retrieved evidence supports or refutes a claim \cite{Thorne:2018:arXiv,Aly:2021:arXiv,Wadden:2020:arXiv}. 

Although large proprietary LLMs are increasingly employed in AFC~\cite{Manakul:2023:EMNLP,Wang:2024:EMNLP,Augenstein:2024:Nature} and deliver strong reasoning and generalization in few-shot settings, their closed nature, high compute and energy costs, and limited reproducibility make them impractical and unsustainable for AFC tasks. Recent evidence shows that smaller transformer models fine-tuned on well-curated fact-checking data can match or even outperform such models~\cite{Setty:2024:SIGIRa}. Open-weight LLMs thus present a more efficient and transparent alternative, offering strong performance when adapted with parameter-efficient methods such as LoRA~\cite{Hu:2021:arXiv,Dettmers:2023:arXiv}. Yet, most AFC systems either rely on few-shot prompting or fine-tune separate models for each subtask~\cite{yoon2025team}, leading to redundant computation, and limited knowledge transfer across related sub-tasks.

To address these limitations, we investigate \textbf{multi-task learning (MTL)} as a unified framework for training open LLMs to perform all three AFC tasks jointly. MTL has long been recognized as a mechanism for improving generalization and sample efficiency through inductive transfer \cite{Zhang:2023:EACL}, and recent instruction-tuned and adapter-based models such as FLAN, T0, and UnifiedQA demonstrate the potential of shared supervision for robust reasoning across tasks \cite{Liu:2019:arXiv,LlamaFactory:2024:arXiv}. Yet, its potential for fact-checking remains largely unexplored, despite the strong semantic dependencies between claim detection, evidence re-ranking, and stance detection. We argue that jointly fine-tuning LLMs on these related tasks can promote representation sharing, stabilize training, and reduce redundancy across the verification pipeline, ultimately yielding a single, parameter-efficient model that balances accuracy and interpretability.

In this work, we study MTL for open-weight LLMs using the Qwen3 family ranging from 0.6B to 8B parameters. Our setup includes two headed transformer variants, classification (CLS) and causal language modeling (CLM), and an instruction-tuned (IT) model using the standard LLM output head. We fine-tune all models jointly with LoRA adapters. Our experiments examine how task-specific loss weighting, task ordering, and model scale influence performance across claim detection, evidence re-ranking, and stance detection. Empirically, MTL outperforms zero-shot, few-shot, and non-LLM baselines and sometimes matches or exceeds single-task fine-tuning, while consolidating the entire fact-checking pipeline into one deployable model. Beyond accuracy, the approach is efficient in computation, memory, and energy use, aligning with sustainable AI principles~\cite{Schwartz:2020:CACM,Strubell:2019:ACL}, while enabling scalable fact-checking systems that support journalists, researchers, and platforms in countering misinformation.

We design our experiments around the following research questions: \textbf{(RQ1)} How does multi-task learning compare with single-task and zero- or few-shot settings across core fact-checking tasks? \textbf{(RQ2)} Which configuration, headed (CLS or CLM) or instruction-tuned (IT), achieves the best balance between accuracy and efficiency? \textbf{(RQ3)} How do task weighting, training order, and model scale affect optimization, transfer, and generalization?  

Our contributions are threefold: (1) a \textbf{unified framework} for jointly fine-tuning open models on claim detection, evidence re-ranking, and stance detection; (2) a \textbf{systematic evaluation} of MTL architectures, loss weighting, and scaling strategies; and (3) empirical findings and practical guidelines showing that multi-task learning yields up to \textbf{44\%, 54\%, and 31\% }relative improvements in Macro-F1 for claim detection, evidence re-ranking, and stance detection, respectively, while promoting transparency and sustainability compared to large proprietary systems. 

We use public datasets and fine-tune using open weight models and provide full implementation with all the experimental settings in the form of config files to ensure faithful reproducibility~\footnote{\url{https://github.com/factiverse/mtl-afc-fine-tune-llms}}
\section{Methodology}
\label{sec:method}
In this section we present a unified MTL framework for fine-tuning an LLM for AFC. The LLM is jointly fine-tuned on three related subtasks: \textbf{(i)} claim detection, \textbf{(ii)} evidence re-ranking, and \textbf{(iii)} stance detection. By sharing a common backbone and optimizing all heads simultaneously, the model learns a coherent representation of factual consistency and argumentation.

We keep the pretrained transformer backbone $\Theta$ frozen and integrate lightweight, trainable QLoRA adapters into selected attention and feed-forward layers. This setup enables parameter-efficient fine-tuning while retaining the model’s general linguistic capabilities. Each task is equipped with its own compact head, either a text classification (CLS) or causal language modeling (CLM) layer and gradients from all tasks update only the adapter and head parameters, encouraging shared learning without catastrophic forgetting.

Let $\Theta$ denote the frozen parameters of a pretrained transformer with $L$ layers. For an input sequence $x = [x_1, \ldots, x_n]$, the layerwise representations are computed as:
\begin{equation}
    h^{(0)} = E(x), \qquad
    h^{(l)} = f_{\Theta}^{(l)}(h^{(l-1)}), \quad l = 1, \ldots, L,
\end{equation}
where $E(\cdot)$ is the embedding function and $f_{\Theta}^{(l)}$ denotes the transformer block at layer $l$.

Each selected layer incorporates a QLoRA adapter:
\begin{equation}
    \tilde{h}^{(l)} = h^{(l)} + B^{(l)} A_r^{(l)} h^{(l)},
\end{equation}
where $A_r^{(l)} \in \mathbb{R}^{r \times d}$ and $B^{(l)} \in \mathbb{R}^{d \times r}$ with $r \ll d$. Only the low-rank parameters $\{A_r^{(l)}, B^{(l)}\}$ and the task-specific heads are updated, while $\Theta$ remains frozen.

\paragraph{Task Classification Heads}
Each task $t \in \mathcal{T} = \{\text{CD}, \text{ER}, \text{SD}\}$ has a small head parameterized by $(W_t, b_t)$, operating on the shared representation $h^{(L)}$.  
For single-input classification (e.g., claim detection),
\begin{equation}
    \hat{y}_t = \text{softmax}(W_t h^{(L)} + b_t), \qquad 
    \mathcal{L}_t = -\sum_i y_i^{(t)} \log \hat{y}_i^{(t)}.
\end{equation}
For re-ranking or stance stance detection with text-pairs,
\begin{equation}
    \hat{y}_t = \text{softmax}(W_t [h_a^{(L)}; h_b^{(L)}] + b_t), \qquad 
    \mathcal{L}_t = -\sum_i y_i^{(t)} \log \hat{y}_i^{(t)}.
\end{equation}
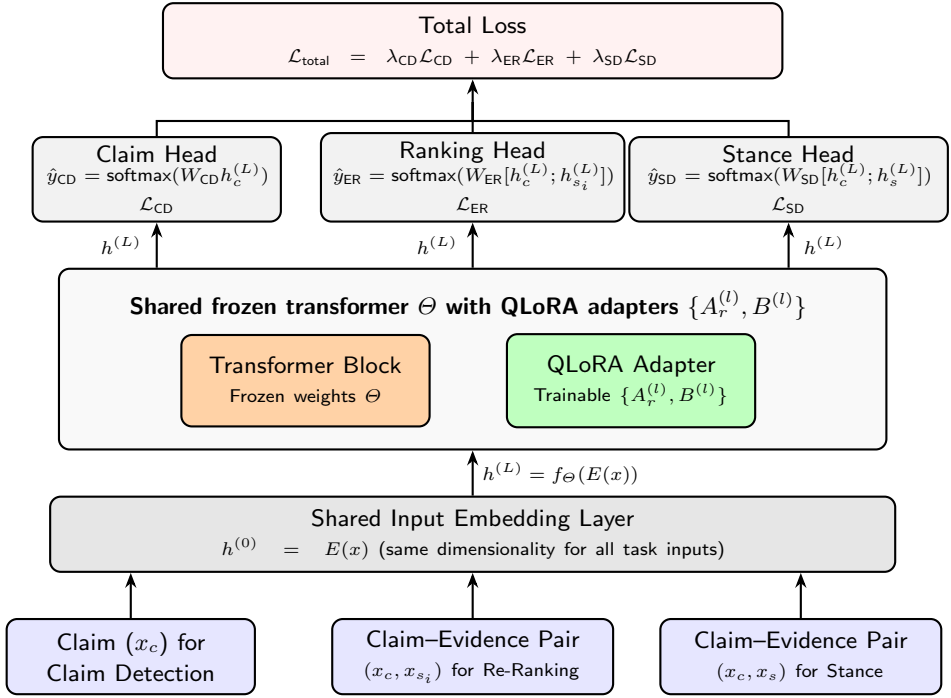
\begin{figure}[ht!]
\centering
\usetikzlibrary{arrows.meta, positioning, calc}

\begin{tikzpicture}[
    font=\sffamily,
    box/.style={draw, thick, rounded corners, align=center, minimum width=3.3cm, minimum height=1cm},
    bigbox/.style={draw, thick, rounded corners, align=center, minimum width=11cm, fill=gray!5, inner ysep=8pt},
    arrow/.style={thick, -{Stealth[length=2mm]}}
]

\node[box, fill=blue!10, text width=3cm, align=center] (claim_in)
{\footnotesize Claim ($x_c$) for Claim Detection};

\node[box, fill=blue!10, text width=3.5cm, align=center, right=1cm of claim_in] (rank_in)
{\footnotesize Claim–Evidence Pair \\[-1pt]\scriptsize $(x_c, x_{s_i})$ for Re-Ranking};

\node[box, fill=blue!10, text width=3.5cm, align=center, right=0.6cm of rank_in] (stance_in)
{\footnotesize Claim–Evidence Pair \\[-1pt]\scriptsize $(x_c, x_s)$ for Stance};

\node[box, fill=gray!20, text width=11cm, above=0.6cm of rank_in, align=center] (embed)
{\footnotesize Shared Input Embedding Layer \\[-1pt]\scriptsize $h^{(0)} = E(x)$ (same dimensionality for all task inputs)};

\draw[arrow] (claim_in.north) -- (embed.south -| claim_in.north);
\draw[arrow] (rank_in.north) -- (embed.south -| rank_in.north);
\draw[arrow] (stance_in.north) -- (embed.south -| stance_in.north);

\node[bigbox, above=0.6cm of embed] (main) {
    \begin{minipage}{9cm}
        \centering
        \textbf{\footnotesize Shared frozen transformer $\Theta$ with QLoRA adapters $\{A_r^{(l)}, B^{(l)}\}$}\\[4pt]
        \begin{tikzpicture}[every node/.style={box}]
            \node[fill=orange!35, text width=3cm, align=center] (tb)
            {\footnotesize Transformer Block\\[-1pt]\scriptsize Frozen weights $\Theta$};
            \node[fill=green!25, text width=3cm, align=center, right=1cm of tb]
            {\footnotesize QLoRA Adapter\\[-1pt]\scriptsize Trainable $\{A_r^{(l)}, B^{(l)}\}$};
        \end{tikzpicture}
    \end{minipage}
};

\draw[arrow] (embed.north) -- node[right, xshift=0pt]{\scriptsize $h^{(L)} = f_{\Theta}(E(x))$} (main.south);

\node[box, fill=gray!10, above=0.6cm of main, xshift=-4.2cm, text width=3cm] (stance)
{\footnotesize Claim Head\\[-1pt]\scriptsize $\hat{y}_{\text{CD}} = \text{softmax}(W_{\text{CD}}h^{(L)}_c)$\\[-1pt]$\mathcal{L}_{\text{CD}}$};

\node[box, fill=gray!10, above=0.6cm of main, text width=3.8cm] (ranking)
{\footnotesize Ranking Head\\[-1pt]\scriptsize $\hat{y}_{\text{ER}} = \text{softmax}(W_{\text{ER}}[h_c^{(L)}; h_{s_i}^{(L)}])$\\[-1pt]$\mathcal{L}_{\text{ER}}$};

\node[box, fill=gray!10, above=0.6cm of main, xshift=4.2cm, text width=4cm] (claim)
{\footnotesize Stance Head\\[-1pt]\scriptsize $\hat{y}_{\text{SD}} = \text{softmax}(W_{\text{SD}}[h_c^{(L)}; h_s^{(L)}])$\\[-1pt]$\mathcal{L}_{\text{SD}}$};

\draw[arrow] (main.north -| stance.south) -- node[left, xshift=-2pt]{\scriptsize $h^{(L)}$} (stance.south);
\draw[arrow] (main.north) -- node[left, xshift=-2pt]{\scriptsize $h^{(L)}$} (ranking.south);
\draw[arrow] (main.north -| claim.south) -- node[right, xshift=2pt]{\scriptsize $h^{(L)}$} (claim.south);

\node[box, fill=pink!20, text width=8cm, above=0.7cm of ranking] (loss)
{\footnotesize Total Loss \\[-1pt]\scriptsize $\mathcal{L}_{\text{total}} = 
\lambda_{\text{CD}}\mathcal{L}_{\text{CD}} +
\lambda_{\text{ER}}\mathcal{L}_{\text{ER}} +
\lambda_{\text{SD}}\mathcal{L}_{\text{SD}}$};

\coordinate (merge) at ([yshift=0.4cm]ranking.north);
\draw[thick] (stance.north) -- ++(0,0.2) -| (merge);
\draw[thick] (ranking.north) -- ++(0,0.2) -| (merge);
\draw[thick] (claim.north) -- ++(0,0.2) -| (merge);
\draw[arrow] (merge) -- (loss.south);

        \vspace{-20pt}
\end{tikzpicture}

\caption{Overview of the multi-task QLoRA fact-checking architecture.}
\label{fig:fact_architecture}
\end{figure}
\paragraph{Causal LM Heads}  
To support generative reasoning, a causal LM head predicts the next token in a sequence:
\begin{equation}
    p(x_t \mid x_{<t}) = \text{softmax}(W_{\text{LM}} h_t^{(L)} + b_{\text{LM}}), \qquad
    \mathcal{L}_{\text{LM}} = -\sum_{t=1}^{n} \log p(x_t \mid x_{<t}).
    \vspace{-5pt}
\end{equation}
\subsection{Multi-Task Learning (MTL) Training}
During multi-task training, each batch $\mathcal{B}$ contains examples from multiple tasks, claim detection $(x_c)$, evidence re-ranking $(x_c, x_{s_i})$, and stance detection $(x_c, x_s)$, which are first embedded into a shared representation $h^{(0)} = E(x)$ and passed through the frozen transformer backbone $\Theta$ augmented with trainable QLoRA adapters $\{A_r^{(l)}, B^{(l)}\}$. The resulting contextual states $h^{(L)}$ are fed into task-specific heads that output logits $\hat{y}_t = W_t h^{(L)} + b_t$, where only active task samples contribute to their corresponding losses $\mathcal{L}_t$ (inactive ones are masked with $y_t = -100$). The total loss for the batch is defined as
\begin{equation}
    \mathcal{L}_{\text{total}}(\mathcal{B}) = \sum_{t \in \mathcal{T}(\mathcal{B})} \lambda_t \, \mathcal{L}_t(\mathcal{B}),
\end{equation}
where $\lambda_t$ controls task importance. Gradients from $\mathcal{L}_{\text{total}}$ are backpropagated through all active heads and shared adapters but not the frozen backbone:
\begin{align}
    \frac{\partial \mathcal{L}_{\text{total}}}{\partial B^{(l)}} 
    &= G^{(l)} (A_r^{(l)} h^{(l)})^{\!\top}, \qquad
    \frac{\partial \mathcal{L}_{\text{total}}}{\partial A_r^{(l)}} 
    = (B^{(l)})^{\!\top} G^{(l)} (h^{(l)})^{\!\top}, 
    \frac{\partial \mathcal{L}_{\text{total}}}{\partial \Theta} &= 0,
\end{align}
where $G^{(l)} = \frac{\partial \mathcal{L}_{\text{total}}}{\partial \tilde{h}^{(l)}}$ denotes the backpropagated gradient at layer $l$. This enables all task heads to jointly optimize shared representations while updating only the lightweight adapter and head parameters, as illustrated in Figure~\ref{fig:fact_architecture}.

\paragraph{Instruction Tuning}

\begin{figure*}[t]
\centering
\small
\setlength{\fboxsep}{6pt}
\fbox{
\begin{minipage}{0.97\linewidth}
\ttfamily
\textbf{Claim detection prompt:}
Analyze the following text and determine if it contains a claim that can be fact-checked.
Respond with checkworthiness label as \{label\_str\}.

\textbf{Evidence ranking prompt:}
Given the following Query (containing a claim and associated question) and the Document
snippet, predict if the snippet contains relevant evidence for the claim or the answer
to the question to predict the veracity of the claim. Respond with \{label\_str\}.

\textbf{Stance detection prompt:}
Evaluate the following claim and determine its veracity. Respond with \{label\_str\} based on available evidence.

Guidelines:
- SUPPORTS: Evidence clearly confirms the claim
- REFUTES: Evidence clearly contradicts the claim
- PARTIALLY SUPPORTS: Evidence mostly supports the claim with minor discrepancies or limitations
- PARTIALLY REFUTES: Evidence contradicts part of the claim while supporting other parts, or shows the claim is mostly but not entirely accurate
\end{minipage}
}
\caption{Prompt used for instruction tuning. For few-shot variations, the prompt is appended with examples from the training data with one example per label.}
\label{fig:prompt}
\end{figure*}

In addition to CLS and CLM fine-tuning, we employ \textbf{instruction tuning} implemented via the Unsloth framework\footnote{\url{https://github.com/unslothai/unsloth}}, where each training sample is formatted as a natural-language \emph{instruction–response} pair. This aligns the model to follow prompts across tasks by optimizing the loss:
\begin{equation}
    \mathcal{L}_{\text{IT}} = -\sum_{t=1}^{n} \log p(y_t \mid y_{<t}, I, x),
\end{equation}
where $I$ denotes the task-specific instruction, $x$ the input context, and $y$ the desired output. Instruction tuning refines the model’s task-following behavior  without introducing additional task-specific heads. The instructions are listed in Figure \ref{fig:prompt}

\section{Experimental Setup}
\label{sec:expsetup}

\subsection{Datasets and Metrics}

In this section we describe the datasets and metrics we use. A full overview of dataset distribution is shown in Table \ref{tab:data_distribution_final}.

\paragraph{Claim detection.}
We use the CheckThat! 2024 English subset~\cite{Barron:2024:CLEF}. This dataset consists of manually annotated statements from the US political debates that are deemed to be either checkworthy (T) or non-checkworhty (F) for fact-checking.

\paragraph{Evidence Re-ranking}
Since no dedicated dataset exists for evidence re-ranking, we adapt the \averitec~dataset\cite{Schlichtkrull:2023:NeurIPS}, which contains real-world claims verified by professional fact-checkers (e.g., Politifact, FullFact). Each claim is paired questions used to query Google retrieving up to 100 web documents. These are segmented into sentences to create snippet-level candidates. Human annotators then answer each question using the retrieved text, and we retain only extractive answers to preserve explicit evidence spans. Snippets containing these spans serve as positives, while all others are treated as negatives for training and evaluation.

\paragraph{Stance Detection}
We use an English dataset collected from professional fact-checkers such as Politifact, Snopes, FullFact etc used in previous research~\cite{Setty:2024:SIGIRa}. This dataset contains four-way classification: Supported (Sup), Partially-Supported (P-Sup), Partially Refuted (P-Ref) and Refuted.

\paragraph{Metrics.}
We report F1 scores for each class, along with macro and weighted averages to capture both balanced and distribution-sensitive performance.

\subsection{Models and Hyperparameters}
All experiments use the following hyperparameters unless explicitly mentioned. We fine-tune Qwen3-4B-Instruct for 5 epochs with a batch size of 32, learning rate of 2e-4, and the paged\_adamw\_32bit optimizer. We enable bfloat16 precision and gradient checkpointing for efficiency. Parameter-efficient tuning uses QLoRA (\textit{r=64}, \textit{$\alpha$=16}) with 4-bit NF4 quantization. All tasks, claim detection, stance detection, and evidence re-ranking, are trained jointly with equal weights ($\lambda$=1:1:1) using mixed data batches without curriculum ordering.

\begin{table}[t!]
\centering
\footnotesize
\caption{\footnotesize Data distribution (\% of total data) across Claim Detection, Evidence Ranking, and Stance Detection tasks and train, validation and test splits.}
\resizebox{\textwidth}{!}{
\begin{tabular}{l|ccr|ccr|ccccr}
\toprule
\textbf{Split} &
\multicolumn{3}{c|}{\textbf{Claim Detection}} &
\multicolumn{3}{c|}{\textbf{Ev. Re-Ranking}} &
\multicolumn{5}{c}{\textbf{Stance Detection}} \\
\cmidrule(lr){2-4} \cmidrule(lr){5-7} \cmidrule(lr){8-12}
 & True & False & Total & Rel & Nrel & Total & Sup & P-Sup & P-Ref & Ref & Total \\
\midrule
\textbf{Training}   & 24.1 & 75.9 & 22,501 & 8.0 & 92.0 & 6,786 & 15.8 & 21.2 & 13.6 & 49.4 & 32,238 \\
\textbf{Validation} & 23.1 & 76.9 & 1,032  & 8.1 & 91.9 & 755   & 16.9 & 20.6 & 17.7 & 44.8 & 1,000  \\
\textbf{Test}       & 34.0 & 66.0 & 318    & 7.9 & 92.1 & 2,783 & 16.4 & 20.4 & 12.7 & 50.6 & 2,566  \\
\bottomrule
\end{tabular}
}
\vspace{-10pt}
\label{tab:data_distribution_final}
\end{table}
\section{Experimental Results}

We evaluate single-task (STL) and multi-task (MTL) learning across claim detection and stance detection. We analyze (i) CLS vs.\ CLM heads, (ii) label masking, (iii) training schedules, and (iv) performance compared to baselines. 

\subsection{Task-Specific Evaluation}
To answer \textbf{RQ1}, for each task, we compare fine-tuned LLMs trained either on single tasks (STL) or on all three tasks jointly (MTL), using the Qwen3-4B model with three configurations: a causal language modeling (CLM) head, a classification head (CLS), and an instruction-tuned (IT) variant. We further include zero-shot and few-shot in-context learning (ICL) baselines. Additionally, we evaluate non-LLM baselines fine-tuned specifically for each task: XLM-RoBERTa-Large for claim detection, stance detection and BGE-reranker~\cite{chen2024bge} (based on XLM-RoBERTa-Large) for evidence retrieval.

\subsubsection{Claim Detection}
As shown in Table~\ref{tab:claim_evidence_combined}, all fine-tuned LLM variants outperform both the zero-shot and few-shot ICL baselines as well as the non-LLM baseline on the claim detection task. The fine-tuned Qwen3-4B model with a classification (CLS) head achieves the highest performance, followed closely by its multi-task learning (MTL) counterpart. The slight reduction in performance for the MTL model can be attributed to the trade-off introduced by jointly fine-tuning the model across all three tasks.

\subsubsection{Evidence Re-ranking}

As shown in Table~\ref{tab:claim_evidence_combined}, all fine-tuned LLM models substantially outperform the zero-shot and few-shot ICL baselines, as well as the non-LLM BGE-based re-rankers, on the evidence re-ranking task. Among the fine-tuned models, the Qwen3-4B model with a classification (CLS) head achieves the best overall performance, with high scores across both relevant and not-relevant classes. Its multi-task learning (MTL) classification variant performs comparably, indicating that multi-task fine-tuning preserves strong ranking ability while adding generalization across related tasks. The instruction-tuned and causal LM variants show moderate performance improvements over baselines but remain below the CLS configuration, suggesting that explicit classification objectives are particularly effective for relevance re-ranking.
\begin{table}[t!!]
\centering
\footnotesize
\caption{\footnotesize Comparison of STL, MTL, and Zero/Few-shot methods on Claim Detection and Evidence Re-Ranking tasks (F1 scores in \%). $*$ For evidence re-ranking bge-reranker-v2-m3~\cite{chen2024bge} is used which is based on XLM-RoBERTa-Large. $^{\dagger}$ indicates statistically significant results ($p\leq0.05$, Bonferroni corrected) }
\label{tab:claim_evidence_combined}
\resizebox{\textwidth}{!}{
\begin{tabular}{lcccccccc}
\toprule
\textbf{Method} & \multicolumn{4}{c}{\textbf{Claim Detection}} & \multicolumn{4}{c}{\textbf{Evidence Re-Ranking}} \\
\cmidrule(lr){2-5} \cmidrule(lr){6-9}
 & \textbf{T-F1} & \textbf{F-F1} & \textbf{Mac-F1} & \textbf{Wei-F1} & \textbf{Rel-F1} & \textbf{NRel-F1} & \textbf{Mac-F1} & \textbf{Wei-F1} \\
\midrule
\multicolumn{9}{l}{\textbf{Single-Task Learning (STL)}} \\
STL Causal LM& 78.18 & 86.51 & 82.35 & 83.68 & 12.82 & 74.41 & 43.61 & 69.52 \\
STL Classification (CLS) & \textbf{89.55}$^{\dagger}$ & \textbf{95.17}$^{\dagger}$ & \textbf{92.36}$^{\dagger}$ & \textbf{93.26}$^{\dagger}$ & \textbf{66.13}$^{\dagger}$ & \textbf{96.66}$^{\dagger}$ & \textbf{81.39}$^{\dagger}$ & \textbf{94.24}$^{\dagger}$ \\
STL Instruction Tuning & 81.30 & 92.51 & 86.92 & 88.70 & 42.40 & 95.80 & 46.10 & 91.60 \\
\midrule
\multicolumn{9}{l}{\textbf{Multi-Task Learning (MTL)}} \\
MTL CLM & 70.58 & 86.11 & 90.71 & 80.83 & 24.82 & 93.87 & 59.35 & 88.39 \\
MTL CLS & 88.37 & 94.06 & 91.21 & 92.12 & 64.15 & 96.63 & 80.39 & 94.05 \\
MTL Instruction Tuning & 85.40 & 93.70 & 89.60 & 90.90 & 42.20 & 94.80 & 68.50 & 90.60 \\
\midrule
\multicolumn{9}{l}{\textbf{Zero-shot / Few-shot Baselines}} \\
Zero-shot & 80.40 & 71.30 & 75.90 & 77.30 & 39.70 & 92.10 & 65.90 & 87.90 \\
Few-shot ICL & 69.30 & 82.80 & 76.10 & 78.20 & 39.70 & 93.20 & 66.50 & 89.00 \\
\midrule
\multicolumn{9}{l}{\textbf{Non-LLM Baselines}} \\
XLM-RoBERTa-Large$^*$ & 80.00 & 90.60 & 85.30 & 86.99 & 43.90 & 92.50 & 68.20 & 84.90 \\
\bottomrule
\vspace{-15pt}
\end{tabular}
}
\end{table}

\begin{table}[ht!!]
\centering
\footnotesize
\caption{\footnotesize Comparison of STL, MTL, and Zero/Few-shot methods for Stance Detection task. Unlike the claim detection and evidence re-ranking stance detection results are not statistically significant. P-Sup/Ref refers to Partially Supported/Refuted.}
\begin{tabular}{lcccccc}
\toprule
\footnotesize
\textbf{Method} & \textbf{Sup-F1} & \textbf{P-Sup-F1} & \textbf{P-Ref-F1} & \textbf{Ref-F1} & \textbf{Macro-F1} & \textbf{Weig-F1} \\
\midrule
\multicolumn{7}{l}{\textbf{Single-Task Learning (STL)}} \\
STL Causal LM& 62.06 & 61.66 & 40.34 & 85.98 & 62.51 & 70.63 \\
STL Classification (CLS) & \textbf{76.33} & 67.78 & 47.36 & 85.97 & \textbf{69.36} & \textbf{75.22} \\
STL Instruction Tuning & 74.90 & \textbf{69.50} & 38.30 & \textbf{86.50} & 67.30 & 75.80 \\
\midrule
\multicolumn{7}{l}{\textbf{Multi-Task Learning (MTL)}} \\
MTL Causal LM (CLM)& 73.97 & 66.54 & \textbf{47.80} & 76.50 & 68.01 & 70.48 \\
MTL Classification & 73.90 & 66.98 & 44.79 & 81.00 & 66.67 & 72.40 \\
MTL Instruction Tuning & 74.80 & 68.20 & 45.40 & 85.60 & 68.50 & 74.60 \\
\midrule
\multicolumn{7}{l}{\textbf{Zero-shot / Few-shot Baselines}} \\
Zero-shot & 66.00 & 25.00 & 29.40 & 80.20 & 50.20 & 62.60 \\
Few-shot ICL & 64.10 & 26.60 & 33.80 & 79.80 & 51.10 & 59.80 \\
\midrule
\multicolumn{7}{l}{\textbf{Non LLM Baseline}}\\
XLM-RoBERTa-Large & 74.80 & 69.30 & 38.10 & 86.20 & 67.10 & 74.34 \\
\bottomrule
\end{tabular}
\vspace{-10pt}
\label{tab:stance}
\end{table}

\subsection{Stance Detection}

As shown in Table~\ref{tab:stance}, all fine-tuned LLM variants outperform the zero-shot and few-shot ICL baselines across all stance labels, demonstrating the clear benefit of task-specific fine-tuning. Among the single-task models, the Qwen3-4B with a classification (CLS) head achieves the highest overall performance, while the instruction-tuned variant performs comparably and yields the best results on several individual stance categories. The multi-task instruction-tuned model also performs strongly, maintaining competitive results across supported, partially supported, and refuted classes, indicating effective transfer learning across related tasks. In contrast, zero-shot and few-shot ICL baselines show substantially lower macro and weighted F1 scores, underscoring that direct fine-tuning remains crucial for nuanced stance detection.

In summary, the multi-task model with a classification (CLS) head consistently outperforms the causal LM and instruction-tuned variants, addressing \textbf{RQ2}. The following sections on loss weighting, task order, and scaling provide further analysis related to \textbf{RQ3}.

\subsection{Effect of Loss Weights}

As defined in Section \ref{sec:method}, we the MTL framework allows setting loss weights specific to the task. The weights are defined as ratios between the tasks. Table~\ref{tab:loss-weights} shows that task weighting impacts multi-task learning performance to some extent. Equal weights (1,1,1) yield balanced but moderate results, while tuned ratios improve cross-task optimization. The (1,4,2) and (4,1,2) configurations achieve the best overall F1 scores, boosting claim detection and stance detection without degrading re-ranking. This suggests that emphasizing claim detection stabilizes training and promotes transfer across subtasks. Overall, informed loss weighting leads to more consistent and robust joint learning compared to uniform optimization. It is also interesting to note that giving higher weight to claim detection and stance detection improves their performance but evidence re-ranking does not benefit from higher weights.

\begin{table}[t!!]
\centering
\footnotesize
\caption{\footnotesize Impact of loss weighting on multi-task learning (MTL)  with classification head (CLS) performance.}
\resizebox{\textwidth}{!}{
\begin{tabular}{
>{\centering\arraybackslash}p{0.8cm}  %
>{\centering\arraybackslash}p{0.8cm}  %
>{\centering\arraybackslash}p{0.8cm}| %
cccccccc
}
\toprule
\multicolumn{3}{c|}{\textbf{Loss Weights}} &
\multicolumn{2}{c}{\textbf{Claim Detection}} &
\multicolumn{2}{c}{\textbf{Evid. Re-Ranking}} &
\multicolumn{2}{c}{\textbf{Stance Detection}} \\
\cmidrule(lr){1-3} \cmidrule(lr){4-5} \cmidrule(lr){6-7} \cmidrule(lr){8-9}
\textbf{C} & \textbf{R} & \textbf{S} &
\textbf{Mac-F1} & \textbf{Wei-F1} &
\textbf{Mac-F1} & \textbf{Wei-F1} &
\textbf{Mac-F1} & \textbf{Wei-F1} \\
\midrule
 1 & 1 & 1 & 91.21 & 92.12 & \textbf{80.39} & \textbf{94.05} & 66.67 & 72.04 \\
 1 & 2 & 4 & 90.55 & 91.51 & 74.39 & 92.69 & 68.51 & 73.49 \\
1 & 4 & 2 & 93.22 & 93.97 & 75.62 & 93.09 & 68.75 & 73.70 \\
2 & 1 & 4 & 91.50 & 92.41 & 74.52 & 92.91 & \textbf{68.86} & \textbf{73.59} \\
4 & 1 & 2 & \textbf{93.99} & \textbf{94.63} & 73.71 & 92.54 & 68.75 & 71.74 \\

\bottomrule
\end{tabular}
}
\vspace{-10pt}
\label{tab:loss-weights}
\end{table}

\subsection{Scaling and Efficiency}
\begin{figure*}[t!!]
\centering
\begin{minipage}[t]{0.48\textwidth}
\centering
\includegraphics[width=\linewidth]{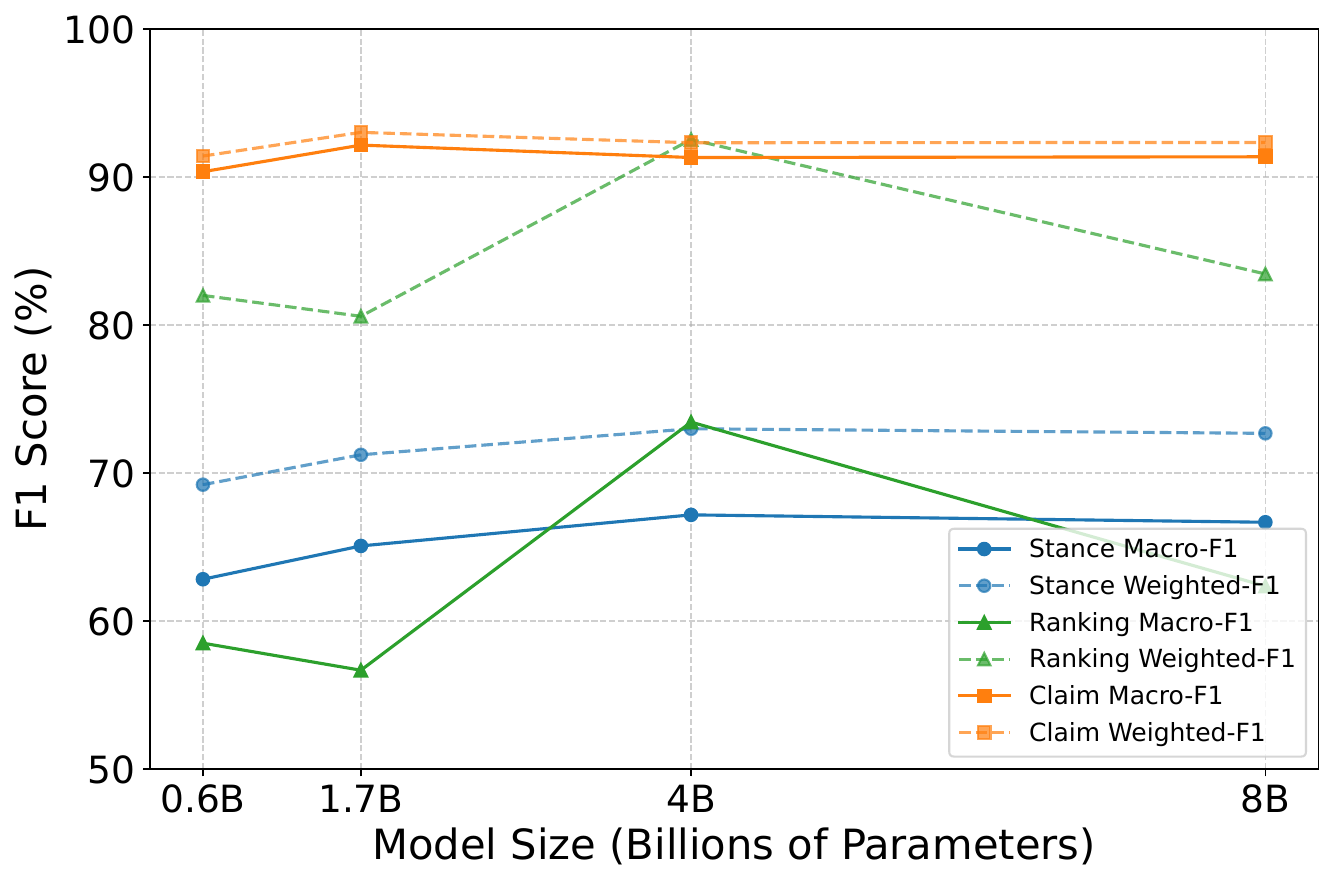}
\caption{Model scaling curve for Qwen3 models of different sizes.}
\label{fig:model_scaling_plot}
\end{minipage}
\hfill
\begin{minipage}[t]{0.48\textwidth}
\centering
\includegraphics[width=\linewidth]{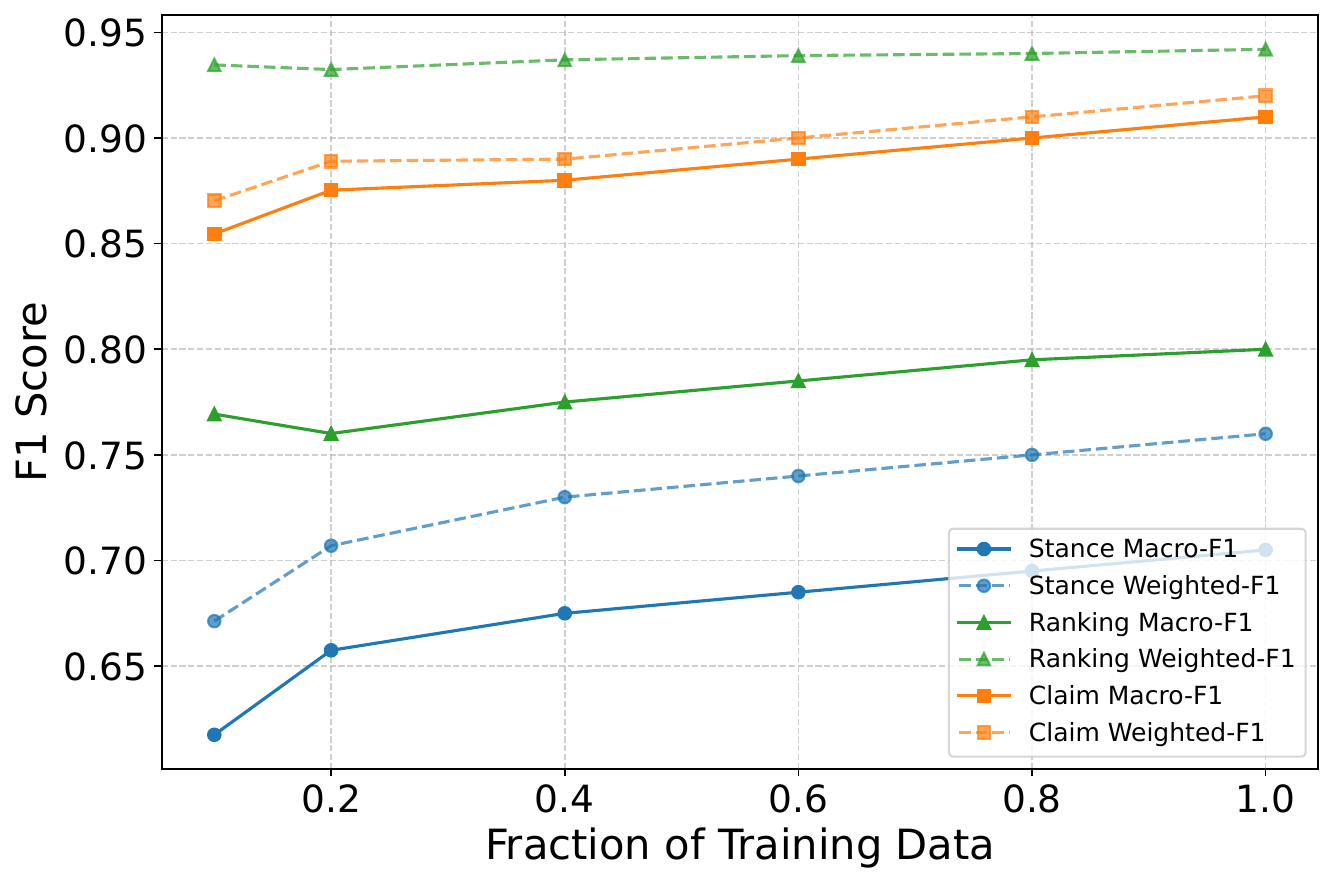}
\caption{Data scaling curve for Qwen3-4b.}
\label{fig:data_scaling_plot}
\vspace{-10pt}
\end{minipage}
\end{figure*}

The model scaling curve in Figure~\ref{fig:model_scaling_plot} shows clear differences across the Qwen3 models. The 4B variant offers the best trade-off between model size and task complexity, performing strongly in evidence re-ranking and stance detection while maintaining stable claim detection. Scaling to 8B brings little additional benefit, suggesting that the architecture reaches saturation in this parameter range. These results indicate that Qwen3 is highly efficient, learning robust factual and reasoning patterns without excessive scaling. We also observed a similar patter with Llama 3.x family of models, due to lack of space we omit those results.

In contrast, the data scaling curve in Figure~\ref{fig:data_scaling_plot} shows steady improvements in Qwen3-4b as training data increases, especially for stance and ranking tasks that depend on richer evidence–claim diversity. Claim detection saturates early since it is a simpler task. This shows that MTL training works even with small fraction of the data but more high-quality training data has a stronger effect on generalization across tasks. 

All models were trained on a single 80GB A100 GPU. Instruction tuning with Unsloth was the fastest configuration (7,624 s, 11.0 samples/s), followed by CLS (26,103 s, 8.5 samples/s) and CLM (45,193 s, 4.2 samples/s). During inference, however, CLS achieved the highest throughput (46 samples/s), with CLM trailing (15.6 samples/s) and prompting remaining the slowest (5.46 samples/s).

\subsection{Impact of Training Schedule}

As shown in Table~\ref{tab:training-order}, the order in which tasks are introduced during multi-task fine-tuning has a measurable impact on overall performance. Among the various scheduling strategies, Claim (C), Evidence re-ranking (R), Stance  detection (S) achieves the highest Macro-F1 (92.28) and Weighted-F1 (93.08) for the claim detection task, indicating that starting with claim-level understanding provides a strong foundation for subsequent evidence and stance modeling. However, the Ranking–Stance–Claim order yields the best results for the evidence ranking task (Macro-F1 82.11, Weighted-F1 94.63) and maintains competitive performance on the other two tasks, suggesting improved cross-task generalization when ranking precedes claim prediction. Overall, both schedules demonstrate that progressive fine-tuning from information retrieval (ranking) to reasoning (stance) and then to classification (claim) leads to balanced, high-performing multi-task models.

\begin{table}[t!!]
\centering
\caption{Effect of training order on multi-task (MTL) with classification head (CLS) performance across claim detection, evidence ranking, and stance detection (F1 scores in \%).}
\begin{tabular}{lcccccc}
\toprule
\textbf{Order} & \multicolumn{2}{c}{\textbf{Claim Detection}} & \multicolumn{2}{c}{\textbf{Evidence Ranking}} & \multicolumn{2}{c}{\textbf{Stance Detection}} \\
\cmidrule(lr){2-3} \cmidrule(lr){4-5} \cmidrule(lr){6-7}
 & \textbf{Mac-F1} & \textbf{Wei-F1} & \textbf{Mac-F1} & \textbf{Wei-F1} & \textbf{Mac-F1} & \textbf{Wei-F1} \\
\midrule
C-S-R & 91.21 & 92.12 & 80.39 & 94.05 & 66.67 & 72.40 \\
C-R-S & \textbf{92.28} & \textbf{93.08} & 78.50 & 93.62 & 66.26 & 71.31 \\
S-R-C & 92.10 & 92.99 & 81.11 & 94.18 & 65.56 & 71.24 \\
S-C-R & 89.76 & 90.84 & 81.54 & 94.27 & 66.27 & 71.81 \\
R-C-S & 91.13 & 92.09 & 81.12 & 94.38 & 65.51 & 70.77 \\
R-S-C & 91.80 & 92.70 & \textbf{82.11} & \textbf{94.63} & \textbf{64.97} & \textbf{71.26} \\
\bottomrule
\end{tabular}
\vspace{-10pt}
\label{tab:training-order}
\end{table}

\section{Analysis and Guidelines}
\label{sec:analysis}

\subsection{Error Analysis}
We perform a qualitative analysis of classification errors for each of the tasks.

\noindent\textbf{Claim Detection.}  Since the label distribution is skewed, all models preferred non-claims over claims. Most errors involved longer, more complex statements, with misclassified claims being 35\% longer and slightly richer in numerical expressions (18.2\% vs.\ 15.9\%), suggesting that length and numerical values drive most misclassifications.

\noindent\textbf{Evidence Re-ranking.}
Re-ranking errors are mostly false negatives, showing that the model struggles to detect implicit relevance between claims and evidence. Many documents provide background or causal context rather than direct verification. For instance, text stating that ``President Macron described Islam as a religion in crisis'' supports a claim about diplomatic fallout but omits the consequence, leading to misclassification. Similar errors occur with paraphrasing, related entities, or temporal shifts, highlighting that AFC demands a broader notion of relevance than standard re-rankers assume.

\noindent\textbf{Stance Detection.}
We manually annotated 100 misclassified stance pairs. Most stance errors arise from confusion between partial and full support (32 cases), followed by long, multi-aspect claims (28) and number-heavy evidence involving dates or percentages (21). A smaller set of errors (19) stem from coreference and ambiguous context, indicating challenges in fine-grained reasoning and discourse interpretation.

\subsection{Guidelines for Practitioners}

Our experiments across model types, loss strategies, and training schedules reveal several key insights for designing effective multi-task fine-tuning pipelines for LLMs in automated fact-checking (AFC):
\begin{description}[leftmargin=0pt, labelsep=1em]
\item[\textbf{Head choice matters.}] Classification (CLS) heads consistently outperform causal language modeling (CLM) heads across all tasks despite being more efficient. CLS objectives provide stronger supervision and more stable optimization for categorical outputs such as claim detection and evidence ranking.

\item[\textbf{Training schedule influences generalization.}] Curricula that progress from retrieval (\textit{re-ranking}) to reasoning (\textit{stance}) to classification (\textit{claim}) yields more balanced performance than random or mixed scheduling.

\item[\textbf{Loss matters to some extent.}] While equal loss weighting performs reasonably, we notice that weighted configurations emphasizing more complex or data-scarce tasks (e.g., stance detection) yield moderate gains.

\item[\textbf{Model scaling shows diminishing returns.}] Qwen3-4B achieves strong performance and parameter efficiency, suggesting that small to mid-sized models can perform competitively under well-tuned MTL.

\item[\textbf{Monitor cross-task trade-offs.}] Improvements on one subtask can lead to mild regressions on others, especially when weighting or order is not tuned.
\end{description}
\section{Related Work}
Automated fact-checking (AFC) research has largely focused on individual sub-tasks in isolation. Early work addressed claim detection~\cite{Hassan:2017:KDD,Elsayed:2019:CLEF} and stance detection~\cite{popat2018declare,Mishra:2019:ICTIR}, while evidence retrieval received limited attention until more recently~\cite{Schlichtkrull:2023:NeurIPS,V:2025:TheWebConf}. Only a few systems have evaluated AFC in an end-to-end setting~\cite{Setty:2024:SIGIR,Hassan:2017:VLDB}, and even these typically rely on separately fine-tuned models for each component.  Over time, benchmarks have evolved from early datasets such as LIAR~\citep{Wang:2017:ACL} and FEVER~\citep{Thorne:2018:arXiv} to more complex multi-hop and multi-evidence datasets, including FEVEROUS~\citep{Aly:2021:arXiv}, HoVer~\citep{Jiang:2021:arXiv}, SciFact-open~\citep{Wadden:2022:EMNLP}, QuanTemp~\citep{Venktesh:2024:SIGIR}, and AVeriTeC~\citep{Schlichtkrull:2023:NeurIPS}. End-to-end AFC benchmarks that combine all three tasks have also emerged~\cite{Setty:2024:SIGIRa,Hassan:2017:VLDB}.  

With the LLMs, their application to AFC has become increasingly common~\cite{Wang:2024:arXiv,Manakul:2023:EMNLP}. More recent studies have explored smaller open-weight models~\cite{Schlichtkrull:2023:NeurIPS,yoon2025team}, while evidence suggests that fine-tuned transformers can still outperform generic LLMs in few-shot reasoning settings~\cite{Setty:2024:SIGIRa}. Multi-task learning (MTL) has been shown to enhance generalization across NLP tasks~\citep{Crawshaw:2020:arXiv,Zhang:2023:EACL}, and prior AFC work has combined related objectives such as rumor detection and stance prediction~\citep{Kochkina:2018:WWW}, AI content detection and misinformation detection~\cite{ren2025llm}, evidence selection with claim classification~\citep{Maillard:2022:arXiv}, entity linking with stance detection~\cite{li-etal-2018-end}, and explanation generation with veracity prediction~\citep{Atanasova:2024:ACL}. However, most of these studies employ encoder-based transformers such as RoBERTa or XLM-R.  

Recently, decoder-only architectures adapted with parameter-efficient methods have demonstrated competitive results~\citep{Zhao:2024:arXiv}. Building on this progress, we evaluate decoder-only models using QLoRA adaptation and compare classification and causal language modeling heads under different task-weighting and scheduling strategies. To our knowledge, this is the first work to jointly fine-tune an open decoder-only model across all major fact-checking components using parameter-efficient adapters and structured multi-task optimization.

\section{Conclusion}
We presented a unified multi-task learning framework for automated fact-checking, jointly fine-tuning open-weight LLMs across claim detection, evidence re-ranking, and stance detection. Using parameter-efficient QLoRA adapters and task-specific heads, our models achieve strong and balanced performance while remaining resource-efficient.  Results show that while multi-task learning does not always surpass single-task models, it performs comparably and clearly outperforms prompting baselines, with substantial gains in evidence re-ranking and consistent improvements across all tasks. The Qwen3-4B model offers the best trade-off between capacity and generalization, confirming the parameter efficiency of modern open LLMs. Overall, this work demonstrates that integrating multiple AFC subtasks within a single, efficient model advances both performance and practicality, paving the way for scalable and transparent fact-checking systems.
\section{Theory of Change}
With the growing use of LLMs, many AFC systems now rely on LLMs for multiple sub-tasks. Current approaches either employ a single large model capable of handling all tasks or use multiple specialized LLMs for different components. Both of these approaches are costly to deploy and maintain. This work explores fine-tuning a single small LLM for all three core AFC tasks. Our experiments show that multi-task learning (MTL) with compact LLMs provide competitive performance while reducing resource requirements, contributing to more \textbf{sustainable AFC systems}. Moreover, enhancing AFC efficiency has \textbf{enormous societal value}, enabling scalable detection of misinformation and disinformation before they can cause harm. We use public datasets, open weight LLMs and release our code fostering further research, use in industry and reproducibility. 
\newpage
\bibliographystyle{plainnat}
\bibliography{conf20yy-xxx.bib}

\begin{thebibliography}{39}
\providecommand{\natexlab}[1]{#1}
\providecommand{\url}[1]{\texttt{#1}}
\expandafter\ifx\csname urlstyle\endcsname\relax
  \providecommand{\doi}[1]{doi: #1}\else
  \providecommand{\doi}{doi: \begingroup \urlstyle{rm}\Url}\fi

\bibitem[Aly et~al.(2021)Aly, Guo, Schlichtkrull, Thorne, Vlachos, Christodoulopoulos, Cocarascu, and Mittal]{Aly:2021:arXiv}
Rami Aly, Zhijiang Guo, Michael Schlichtkrull, James Thorne, Andreas Vlachos, Christos Christodoulopoulos, Oana Cocarascu, and Arpit Mittal.
\newblock The fact extraction and verification over unstructured and structured information (feverous) shared task.
\newblock In \emph{Proceedings of the Fourth Workshop on Fact Extraction and VERification (FEVER) at EMNLP}, volume 2021, pages 1--13, 2021.

\bibitem[Atanasova and Augenstein(2024)]{Atanasova:2024:ACL}
Pepa Atanasova and Isabelle Augenstein.
\newblock Generating explanations for veracity prediction with llms.
\newblock In \emph{ACL}, ACL, 2024.

\bibitem[Augenstein et~al.(2024)Augenstein, Baldwin, Cha, Chakraborty, Ciampaglia, Corney, DiResta, Ferrara, Hale, Halevy, Hovy, Ji, Menczer, Miguez, Nakov, Scheufele, Sharma, and Zagni]{Augenstein:2024:Nature}
Isabelle Augenstein, Timothy Baldwin, Meeyoung Cha, Tanmoy Chakraborty, Giovanni~Luca Ciampaglia, David Corney, Renee DiResta, Emilio Ferrara, Scott Hale, Alon Halevy, Eduard Hovy, Heng Ji, Filippo Menczer, Ruben Miguez, Preslav Nakov, Dietram Scheufele, Shivam Sharma, and Giovanni Zagni.
\newblock Factuality challenges in the era of large language models and opportunities for fact-checking.
\newblock \emph{Nature Machine Intelligence}, 6\penalty0 (8):\penalty0 852--863, 2024.

\bibitem[Barr\'{o}n-Cede\~{n}o et~al.(2020)Barr\'{o}n-Cede\~{n}o, Elsayed, Nakov, Da~San~Martino, Hasanain, Suwaileh, Haouari, Babulkov, Hamdan, Nikolov, Shaar, and Ali]{BarronCedeno:2020:arXiv}
Alberto Barr\'{o}n-Cede\~{n}o, Tamer Elsayed, Preslav Nakov, Giovanni Da~San~Martino, Maram Hasanain, Reem Suwaileh, Fatima Haouari, Nikolay Babulkov, Bayan Hamdan, Alex Nikolov, Shaden Shaar, and Zien~Sheikh Ali.
\newblock Overview of checkthat! 2020: Automatic identification and verification of claims in social media.
\newblock In \emph{International Conference of the Cross-Language Evaluation Forum for European Languages}, pages 215--236. Springer, 2020.

\bibitem[Barr{\'o}n-Cede{\~n}o et~al.(2024)Barr{\'o}n-Cede{\~n}o, Alam, Stru{\ss}, Nakov, Chakraborty, Elsayed, Przyby{\l}a, Caselli, Martino, Haouari, Hasanain, Li, Piskorski, Ruggeri, Song, and Suwaileh]{Barron:2024:CLEF}
Alberto Barr{\'o}n-Cede{\~n}o, Firoj Alam, Julia~Maria Stru{\ss}, Preslav Nakov, Tanmoy Chakraborty, Tamer Elsayed, Piotr Przyby{\l}a, Tommaso Caselli, Giovanni Da~San Martino, Fatima Haouari, Maram Hasanain, Chengkai Li, Jakub Piskorski, Federico Ruggeri, Xingyi Song, and Reem Suwaileh.
\newblock Overview of the clef-2024 checkthat! lab: check-worthiness, subjectivity, persuasion, roles, authorities, and adversarial robustness.
\newblock In \emph{International Conference of the Cross-Language Evaluation Forum for European Languages}, pages 28--52. Springer, 2024.

\bibitem[Chen et~al.(2024)Chen, Xiao, Zhang, Luo, Lian, and Liu]{chen2024bge}
Jianlv Chen, Shitao Xiao, Peitian Zhang, Kun Luo, Defu Lian, and Zheng Liu.
\newblock Bge m3-embedding: Multi-lingual, multi-functionality, multi-granularity text embeddings through self-knowledge distillation, 2024.

\bibitem[Crawshaw(2020)]{Crawshaw:2020:arXiv}
Michael Crawshaw.
\newblock Multi-task learning with deep neural networks: A survey, 2020.

\bibitem[Dettmers et~al.(2023)Dettmers, Pagnoni, Holtzman, and Zettlemoyer]{Dettmers:2023:arXiv}
Tim Dettmers, Artidoro Pagnoni, Ari Holtzman, and Luke Zettlemoyer.
\newblock Qlora: Efficient finetuning of quantized llms, 2023.

\bibitem[Elsayed et~al.(2019)Elsayed, Nakov, Barr{\'o}n-Cedeno, Hasanain, Suwaileh, Da~San~Martino, and Atanasova]{Elsayed:2019:CLEF}
Tamer Elsayed, Preslav Nakov, Alberto Barr{\'o}n-Cedeno, Maram Hasanain, Reem Suwaileh, Giovanni Da~San~Martino, and Pepa Atanasova.
\newblock Overview of the clef-2019 checkthat! lab: automatic identification and verification of claims.
\newblock In \emph{International conference of the cross-language evaluation forum for European languages}, pages 301--321. Springer, 2019.

\bibitem[Guo et~al.(2022)Guo, Schlichtkrull, and Vlachos]{Guo:2022:TACL}
Zhijiang Guo, Michael Schlichtkrull, and Andreas Vlachos.
\newblock A survey on automated fact-checking.
\newblock \emph{Transactions of the Association for Computational Linguistics}, 10:\penalty0 178--206, 2022.

\bibitem[Hassan et~al.(2017{\natexlab{a}})Hassan, Arslan, Li, and Tremayne]{Hassan:2017:KDD}
Naeemul Hassan, Fatma Arslan, Chengkai Li, and Mark Tremayne.
\newblock Toward automated fact-checking: Detecting check-worthy factual claims by claimbuster.
\newblock In \emph{Proceedings of the 23rd ACM SIGKDD International Conference on Knowledge Discovery and Data Mining}, KDD, pages 1803--1812, 2017{\natexlab{a}}.

\bibitem[Hassan et~al.(2017{\natexlab{b}})Hassan, Zhang, Arslan, Caraballo, Jimenez, Gawsane, Hasan, Joseph, Kulkarni, Nayak, Sable, Li, and Tremayne]{Hassan:2017:VLDB}
Naeemul Hassan, Gensheng Zhang, Fatma Arslan, Josue Caraballo, Damian Jimenez, Siddhant Gawsane, Shohedul Hasan, Minumol Joseph, Aaditya Kulkarni, Anil~Kumar Nayak, Vikas Sable, Chengkai Li, and Mark Tremayne.
\newblock Claimbuster: The first-ever end-to-end fact-checking system.
\newblock \emph{Proceedings of the VLDB Endowment}, 10\penalty0 (12):\penalty0 1945--1948, 2017{\natexlab{b}}.

\bibitem[Hu et~al.(2021)Hu, Shen, Wallis, Allen{-}Zhu, Li, Wang, Wang, and Chen]{Hu:2021:arXiv}
Edward~J. Hu, Yelong Shen, Phillip Wallis, Zeyuan Allen{-}Zhu, Yuanzhi Li, Shean Wang, Lu~Wang, and Weizhu Chen.
\newblock Lora: Low-rank adaptation of large language models, 2021.

\bibitem[Jiang et~al.(2020)Jiang, Bordia, Zhong, Dognin, Singh, and Bansal]{Jiang:2021:arXiv}
Yichen Jiang, Shikha Bordia, Zheng Zhong, Charles Dognin, Maneesh Singh, and Mohit Bansal.
\newblock Hover: A dataset for many-hop fact extraction and claim verification.
\newblock In \emph{Findings of the Association for Computational Linguistics: EMNLP 2020}, pages 3441--3460, 2020.

\bibitem[Kochkina et~al.(2018)Kochkina, Liakata, and Augenstein]{Kochkina:2018:WWW}
Elena Kochkina, Maria Liakata, and Isabelle Augenstein.
\newblock All-in-one: Multi-task learning for rumour verification.
\newblock In \emph{WWW}, WWW, pages 236--247, 2018.

\bibitem[Li et~al.(2018)Li, Zhao, Cheng, and Yang]{li-etal-2018-end}
Sizhen Li, Shuai Zhao, Bo~Cheng, and Hao Yang.
\newblock An end-to-end multi-task learning model for fact checking.
\newblock In James Thorne, Andreas Vlachos, Oana Cocarascu, Christos Christodoulopoulos, and Arpit Mittal, editors, \emph{Proceedings of the First Workshop on Fact Extraction and {VER}ification ({FEVER})}, pages 138--144, Brussels, Belgium, November 2018. Association for Computational Linguistics.
\newblock \doi{10.18653/v1/W18-5523}.
\newblock URL \url{https://aclanthology.org/W18-5523/}.

\bibitem[Liu et~al.(2019)Liu, He, Chen, and Gao]{Liu:2019:arXiv}
Xiaodong Liu, Pengcheng He, Weizhu Chen, and Jianfeng Gao.
\newblock Multi-task deep neural networks for natural language understanding, 2019.

\bibitem[Maillard et~al.(2022)Maillard, Minervini, Korhonen, Riedel, and Stenetorp]{Maillard:2022:arXiv}
Jean Maillard, Pasquale Minervini, Anna Korhonen, Sebastian Riedel, and Pontus Stenetorp.
\newblock Multi-task evidence retrieval for question answering and claim verification, 2022.

\bibitem[Manakul et~al.(2023)Manakul, Liusie, and Gales]{Manakul:2023:EMNLP}
Potsawee Manakul, Adian Liusie, and Mark Gales.
\newblock Selfcheckgpt: Zero-resource black-box hallucination detection for generative large language models.
\newblock In \emph{Proceedings of the 2023 Conference on Empirical Methods in Natural Language Processing}, pages 9004--9017, 2023.

\bibitem[Mishra and Setty(2019)]{Mishra:2019:ICTIR}
Rahul Mishra and Vinay Setty.
\newblock Sadhan: Hierarchical attention networks to learn latent aspect embeddings for fake news detection.
\newblock In \emph{Proceedings of the 2019 ACM SIGIR international conference on theory of information retrieval}, pages 197--204, 2019.

\bibitem[Popat et~al.(2018)Popat, Mukherjee, Yates, and Weikum]{popat2018declare}
Kashyap Popat, Subhabrata Mukherjee, Andrew Yates, and Gerhard Weikum.
\newblock Declare: Debunking fake news and false claims using evidence-aware deep learning.
\newblock In \emph{Proceedings of the 2018 Conference on Empirical Methods in Natural Language Processing}, pages 22--32, 2018.

\bibitem[Ren et~al.(2025)Ren, Jiang, Huang, Yang, and Xie]{ren2025llm}
Gang Ren, Li~Jiang, Tingting Huang, Ying Yang, and Ruida Xie.
\newblock Llm-enhanced multi-task joint learning model for misinformation detection.
\newblock \emph{Information Processing \& Management}, 62\penalty0 (6):\penalty0 104305, 2025.

\bibitem[Schlichtkrull et~al.(2023)Schlichtkrull, Guo, and Vlachos]{Schlichtkrull:2023:NeurIPS}
Michael Schlichtkrull, Zhijiang Guo, and Andreas Vlachos.
\newblock Averitec: A dataset for real-world claim verification with evidence from the web.
\newblock \emph{Advances in Neural Information Processing Systems}, 36:\penalty0 65128--65167, 2023.

\bibitem[Schwartz et~al.(2020)Schwartz, Dodge, Smith, and Etzioni]{Schwartz:2020:CACM}
Roy Schwartz, Jesse Dodge, Noah~A. Smith, and Oren Etzioni.
\newblock Green ai.
\newblock \emph{Communications of the ACM}, 63\penalty0 (12):\penalty0 54--63, 2020.

\bibitem[Setty(2024{\natexlab{a}})]{Setty:2024:SIGIR}
Vinay Setty.
\newblock Factcheck editor: Multilingual text editor with end-to-end fact-checking.
\newblock In \emph{Proceedings of the 47th International ACM SIGIR Conference on Research and Development in Information Retrieval}, pages 2744--2748, 2024{\natexlab{a}}.

\bibitem[Setty(2024{\natexlab{b}})]{Setty:2024:SIGIRa}
Vinay Setty.
\newblock Surprising efficacy of fine-tuned transformers for fact-checking over larger language models.
\newblock In \emph{Proceedings of the 47th International ACM SIGIR Conference on Research and Development in Information Retrieval}, pages 2842--2846, 2024{\natexlab{b}}.

\bibitem[Strubell et~al.(2019)Strubell, Ganesh, and McCallum]{Strubell:2019:ACL}
Emma Strubell, Ananya Ganesh, and Andrew McCallum.
\newblock Energy and policy considerations for deep learning in nlp.
\newblock In \emph{Proceedings of the 57th Annual Meeting of the Association for Computational Linguistics}, ACL, pages 3645--3650, 2019.

\bibitem[Thorne et~al.(2018)Thorne, Vlachos, Cocarascu, Christodoulopoulos, and Mittal]{Thorne:2018:arXiv}
James Thorne, Andreas Vlachos, Oana Cocarascu, Christos Christodoulopoulos, and Arpit Mittal.
\newblock Fever: a large-scale dataset for fact extraction and verification, 2018.

\bibitem[V and Setty(2025)]{V:2025:TheWebConf}
Venktesh V and Vinay Setty.
\newblock Factir: A real-world zero-shot open-domain retrieval benchmark for fact-checking.
\newblock In \emph{Companion Proceedings of the ACM on Web Conference 2025}, WWW '25, page 809–812, 2025.
\newblock ISBN 9798400713316.

\bibitem[Venktesh et~al.(2024)Venktesh, Anand, Anand, and Setty]{Venktesh:2024:SIGIR}
V.~Venktesh, Abhijit Anand, Avishek Anand, and Vinay Setty.
\newblock Quantemp: A real-world open-domain benchmark for fact-checking numerical claims.
\newblock In \emph{Proceedings of the 47th International ACM SIGIR Conference on Research and Development in Information Retrieval}, SIGIR '24, pages 650--660, 2024.

\bibitem[Wadden et~al.(2020)Wadden, Lin, Lo, Wang, van Zuylen, Cohan, and Hajishirzi]{Wadden:2020:arXiv}
David Wadden, Shanchuan Lin, Kyle Lo, Lucy~Lu Wang, Madeleine van Zuylen, Arman Cohan, and Hannaneh Hajishirzi.
\newblock Fact or fiction: Verifying scientific claims.
\newblock In \emph{Proceedings of the 2020 Conference on Empirical Methods in Natural Language Processing (EMNLP)}, pages 7534--7550, 2020.

\bibitem[Wadden et~al.(2022)Wadden, Lo, Kuehl, Cohan, Beltagy, Wang, and Hajishirzi]{Wadden:2022:EMNLP}
David Wadden, Kyle Lo, Bailey Kuehl, Arman Cohan, Iz~Beltagy, Lucy~Lu Wang, and Hannaneh Hajishirzi.
\newblock Scifact-open: Towards open-domain scientific claim verification.
\newblock In \emph{Findings of the Association for Computational Linguistics: EMNLP 2022}, pages 4719--4734, 2022.

\bibitem[Wang(2017)]{Wang:2017:ACL}
William~Yang Wang.
\newblock “liar, liar pants on fire”: A new benchmark dataset for fake news detection.
\newblock In \emph{ACL}, ACL, pages 422--426, 2017.

\bibitem[Wang et~al.(2024{\natexlab{a}})Wang, Gangi~Reddy, Mujahid, Arora, Rubashevskii, Geng, Mohammed~Afzal, Pan, Borenstein, Pillai, Augenstein, Gurevych, and Nakov]{Wang:2024:EMNLP}
Yuxia Wang, Revanth Gangi~Reddy, Zain~Muhammad Mujahid, Arnav Arora, Aleksandr Rubashevskii, Jiahui Geng, Osama Mohammed~Afzal, Liangming Pan, Nadav Borenstein, Aditya Pillai, Isabelle Augenstein, Iryna Gurevych, and Preslav Nakov.
\newblock Factcheck-bench: Fine-grained evaluation benchmark for automatic fact-checkers.
\newblock In \emph{Findings of the Association for Computational Linguistics: EMNLP 2024}, pages 14199--14230, November 2024{\natexlab{a}}.

\bibitem[Wang et~al.(2024{\natexlab{b}})Wang, Gangi~Reddy, Mujahid, Arora, Rubashevskii, Geng, Mohammed~Afzal, Pan, Borenstein, Pillai, Augenstein, Gurevych, and Nakov]{Wang:2024:arXiv}
Yuxia Wang, Revanth Gangi~Reddy, Zain~Muhammad Mujahid, Arnav Arora, Aleksandr Rubashevskii, Jiahui Geng, Osama Mohammed~Afzal, Liangming Pan, Nadav Borenstein, Aditya Pillai, Isabelle Augenstein, Iryna Gurevych, and Preslav Nakov.
\newblock Factcheck-bench: Fine-grained evaluation benchmark for automatic fact-checkers.
\newblock In Yaser Al-Onaizan, Mohit Bansal, and Yun-Nung Chen, editors, \emph{Findings of the Association for Computational Linguistics: EMNLP 2024}, pages 14199--14230, Miami, Florida, USA, November 2024{\natexlab{b}}. Association for Computational Linguistics.
\newblock \doi{10.18653/v1/2024.findings-emnlp.830}.
\newblock URL \url{https://aclanthology.org/2024.findings-emnlp.830/}.

\bibitem[Yoon et~al.(2025)Yoon, Jung, Yoon, and Park]{yoon2025team}
Yejun Yoon, Jaeyoon Jung, Seunghyun Yoon, and Kunwoo Park.
\newblock Team humane at averitec 2025: Hero 2 for efficient fact verification.
\newblock In \emph{The Eighth Fact Extraction and VERification Workshop}, page 224, 2025.

\bibitem[Zhang et~al.(2023)Zhang, Yu, Yu, Guo, and Jiang]{Zhang:2023:EACL}
Zhihan Zhang, Wenhao Yu, Mengxia Yu, Zhichun Guo, and Meng Jiang.
\newblock A survey of multi-task learning in natural language processing: Regarding task relatedness and training methods.
\newblock In \emph{Proceedings of the 17th Conference of the European Chapter of the Association for Computational Linguistics}, pages 943--956, May 2023.

\bibitem[Zhao et~al.(2024)Zhao, Chen, Zhang, and Yang]{Zhao:2024:arXiv}
Hang Zhao, Qile~P Chen, Yijing~Barry Zhang, and Gang Yang.
\newblock Advancing single and multi-task text classification through large language model fine-tuning, 2024.

\bibitem[Zheng et~al.(2024)Zheng, Zhang, Zhang, YeYanhan, and Luo]{LlamaFactory:2024:arXiv}
Yaowei Zheng, Richong Zhang, Junhao Zhang, YeYanhan YeYanhan, and Zheyan Luo.
\newblock Llamafactory: Unified efficient fine-tuning of 100+ language models.
\newblock In \emph{Proceedings of the 62nd Annual Meeting of the Association for Computational Linguistics (Volume 3: System Demonstrations)}, pages 400--410, 2024.

\end{thebibliography}

\end{document}